# LM-DSB: A Lightweight Multi-agent System for Automated Daily Safety Bulletin Generation in Construction

**Xianxiang Zhao[1], Advik Mehta[3], Falak Sethi[2], Brian Gue[3], Qipei Mei[4], Lingzi Wu[1]**

[1]Department of Construction Management, University of Washington, United States
[2]Department of Computer Science, University of Alberta, Canada
[3]PCL Industrial Management Inc, Canada
[4]Department of Civil and Environmental Engineering, University of Alberta, Canada

seanzhao@uw.edu, advik@ualberta.ca, fsethi@ualberta.ca, bmgue@pcl.com, qipei.mei@ualberta.ca, lingwu@uw.edu

## 1 -Abstract

Construction safety management continues to face challenges in generating daily safety bulletins (DSBs)—structured daily communications that identify site-specific hazards, communicate safety protocols, and establish preventive measures for workers. Despite their critical role in proactive safety management, manual DSB creation requires 2-3 hours daily and produces inconsistent documentation quality, with empirical evidence showing that 25% of crews never conduct pretask planning at all (Olson et al., 2016). This labor-intensive process diverts safety personnel from field supervision while creating gaps in safety coverage across construction sites. This study proposes the Lightweight Multi-Agent DSB System (LM-DSB), a novel framework integrating retrieval-augmented generation (RAG) with a locally deployable Large Language Model (LLM) to automate DSB generation on consumer-grade personal computers. LM-DSB employs a hierarchical architecture coordinating specialized agents—planner, researcher, and writer—enhanced with domain-specific fine-tuning on construction safety documentation and RAG-based knowledge grounding across 37 health, safety, and environment (HSE) documents. A two-phase validation strategy evaluated system feasibility through comparative quantitative analysis and expert assessment. Phase 1 comparative evaluation demonstrated that the multi-agent RAG system (Llama 3-8B) achieved substantial improvements over all alternative configurations, with strong lexical alignment and content coherence (TF-IDF: 0.70, BLEU: 0.39, ROUGE-1: 0.66) and near-complete document generation (length ratio: 1.01 vs. 0.39 for single-agent RAG). Phase 2 expert evaluation (N=40) confirmed practical acceptability with robust ratings across quality dimensions (mean score = 3.50–4.20 on a 5-point scale), particularly in regulatory compliance (4.20) and factual accuracy (4.10). The findings demonstrate that architectural innovation enables lightweight models to handle safety-critical documentation tasks on consumer-grade hardware, advancing technology democratization in construction safety management. The study acknowledges limitations in knowledge base scope, language constraints, and field deployment validation, suggesting future directions in multimodal integration and seamless workflow adoption.

## 2 Introduction

Construction safety management remains a critical challenge in the Architecture, Engineering, and Construction (AEC) industry, with the sector continuing to report disproportionately high occupational health and safety incident rates despite various risk mitigation strategies (Waehrer et al., 2007). According to the Bureau of Labor Statistics' National Census of Fatal Occupational Injuries (2024), the construction industry recorded 1,075 fatalities, representing approximately 20% of all workplace deaths in the United States. With a fatal injury rate of 9.6 deaths per 100,000 full-time equivalent workers—nearly triple the national average—construction remains one of America's most hazardous occupations (BLS, 2024). To address these persistent risks, the industry has established proactive communication practices as a cornerstone of effective safety management. Among these, daily safety bulletins (DSBs) serve as critical tools for identifying site-specific hazards, communicating required safety protocols, and implementing preventive measures that directly impact worker safety performance (Hassan & Le, 2020).

Daily Safety Bulletins (DSBs), sometimes referred to as daily safety briefs or toolbox talks in construction practice (Kearney et al., 2025; Lee et al., 2025), are largely prepared manually and are highly superintendent-dependent. The substantial time and administrative effort required for preparing DSBs often necessitate reliance on individual judgment and available time, resulting in inconsistent safety coverage and variability in safety communication and documentation quality across sites. Moreover, communication suffers from objective issues, such as omissions, format inconsistencies, and ambiguous terminology, as well as subjective issues like varied interpretation and unclear expression. While project management platforms (e.g., Procore, PlanGrid) and BIM-based documentation systems have improved broader project documentation practices, they fail to meet the specialized requirements of DSBs, which demand accurate extraction of safety-critical information from diverse sources and its transformation into actionable daily guidance.

Despite growing interest in applying natural language processing (NLP) and large language models (LLMs) to construction safety, existing research has not yet demonstrated an effective approach for transforming unstructured safety documentation into automated, professional-grade content generation systems. Current studies either focus on document classification and analysis without addressing generation, or rely on expensive commercial models (≥70B parameters) inaccessible to most construction firms. Two specific gaps remain underexplored: first, the automated generation of daily safety communications—as distinct from document classification, compliance checking, and incident analysis—has received limited research attention despite its practical significance; and second, the respective contributions of enabling technologies such as fine-tuning, retrieval augmentation, and multi-agent orchestration have not been empirically evaluated within a unified system, particularly under the deployment constraints typical of construction organizations.

To address the practical challenges and the research gaps, this study proposes a novel construction safety management framework that leverages large language model (LLM) enhanced with retrieval-augmented generation (RAG) and orchestrated through a multi-agent system. The framework is designed to transform the current labor-intensive, error-prone manual process into a streamlined and responsive workflow for DSB creation. By bridging the gap between unstructured safety documentation and actionable site-level communication, the system reduces preparation time, ensures consistent coverage, and improves the factual accuracy of safety guidance.

The primary contribution of this paper is the development and validation of an early empirical investigation of automated DSB generation using a lightweight (8B parameter) multi-agent system, providing controlled evidence on the architectural requirements for professional-quality safety documentation on consumer-grade hardware. Through comparative experiments against single-agent and manual approaches, the study demonstrates measurable improvements in efficiency, reliability, and adaptability, thereby advancing both the theory and practice of automated construction safety communication.

The remainder of this paper is organized as follows: Section 3 reviews prior work in automated construction safety documentation and NLP applications in construction, identifying key research gaps. Section 4 presents the system development, including model selection, data preparation, and multi-agent architecture. Section 5 evaluates system performance through experimental testing. Section 6 discusses broader implications, and Section 7 concludes with contributions and directions for future research.

## 3 Related Work

This section examines research streams that inform automated DSB systems in construction. As this specialized area exists within broader frameworks, we review two complementary domains: automated construction safety document checking and generation (Section 3.1), and NLP applications in construction including recent advances in retrieval augmentation and multi-agent architectures (Section 3.2). Section 3.3 then synthesizes these findings to identify research gaps and introduce the proposed system.

### *3.1 Automated Construction Safety Document Checking and Generation*

The automation of construction safety documentation has evolved through three distinct phases, each advancing analytical and monitoring capabilities while leaving the generation of daily safety communications unaddressed.

Phase 1: Rule-Based Expert Systems (1990s-2010s). Early automation efforts employed expert rules and fuzzy logic to encode safety knowledge into computational frameworks (Russell & Fayek, 1994; Elzarka et al., 1999). While these systems achieved systematic knowledge representation and automated activity-safety associations, they remained constrained by rigid predefined logic incapable of adapting to dynamic site conditions—a fundamental architectural limitation rather than an implementation issue requiring incremental refinement.

Phase 2: Data-Driven Perception Systems (2010s-2020). Research shifted toward real-time monitoring through BIM integration (Ganah & John, 2015; Marefat et al., 2019), IoT sensors (Teizer et al., 2010; Martinez et al., 2020), and AI-driven risk prediction (Ayhan & Tokdemir, 2020; Peng et al., 2023), representing Li's (2019) second-generation intelligent hardware and data collection paradigm. Despite achieving unprecedented capabilities in hazard detection and predictive analytics, these systems focused exclusively on perception and analysis rather than generation—leaving the critical task of transforming monitoring insights into actionable daily safety communications fundamentally unaddressed.

Phase 3: AI-Driven Generative Systems (2020-Present). Recent applications of LLMs demonstrate capabilities in natural language generation substantially exceeding rule-based predecessors (Baek et al., 2025). However, existing implementations focus on information retrieval, compliance checking, and post-hoc incident analysis rather than proactive daily communication generation. Moreover, these systems rely predominantly on commercial cloud-based APIs, and no study has systematically evaluated which combination of enabling technologies—fine-tuning, retrieval augmentation, or architectural design—is necessary to produce safety documents of acceptable quality.

This three-phase evolution reveals a persistent pattern: despite progressively sophisticated capabilities in safety monitoring and analysis, each phase has focused on perception, classification, or compliance verification rather than the proactive generation of daily safety communications—documents that must synthesize dynamic site conditions, regulatory requirements, and environmental factors into actionable guidance.

### *3.2 Natural Language Processing for Construction*

NLP applications in construction have evolved from document analysis toward generation, providing the technological building blocks for automated safety communication. We examine this evolution across three dimensions: document understanding capabilities, generative model emergence, and architectural innovations.

Document Classification and Analysis (Early NLP). Initial NLP applications in construction focused on document understanding through traditional machine learning approaches including SVMs and k-NNs for classification (Caldas et al., 2002; Caldas & Soibelman, 2003), construction-specific ontologies for terminology extraction (El-Diraby & Zhang, 2006), and advanced models for safety incident analysis (Ganguli et al., 2021; Tixier et al., 2016). However, these approaches addressed only Natural Language Understanding—analyzing existing documents rather than generating new safety content.

LLMs and Retrieval Augmentation (Recent NLP).

The emergence of LLMs introduced generative capabilities through pre-training on vast corpora, demonstrating applications in construction scheduling (Prieto et al., 2023), risk management(Aladağ, 2023), and incident analysis (Smetana et al., 2024). To address reliability concerns, researchers enhanced LLMs with retrieval-augmented generation (RAG), incorporating external knowledge bases for improved factual accuracy (Chang et al., 2024; Izacard et al., 2022; Uhm et al., 2025). However, existing RAG implementations in construction focus on question-answering and compliance verification against static knowledge bases, rather than synthesizing multiple dynamic inputs—site conditions, weather, regulatory codes, and project context—into coherent daily safety communications.

Multi-Agent Architectures (Architectural Innovation). Recent computer science research addresses single-model limitations through multi-agent systems where specialized agents collaboratively solve complex tasks (Wang & Duan, 2024), with NLP applications demonstrating that task decomposition across coordinated agents can compensate for individual model constraints (Hong et al., 2024; G. Li et al., 2023). However, multi-agent architectures have not been applied to construction safety documentation, and no study has empirically isolated the contribution of agent orchestration relative to other system components—such as retrieval augmentation or domain-specific fine-tuning—within a controlled experimental design.

In summary, NLP evolution in construction has produced increasingly capable technologies for safety-related tasks, yet these advances remain fragmented: LLMs have been applied to analysis and retrieval, RAG to static knowledge grounding, and multi-agent architectures to general-domain coordination. How these components may be integrated for proactive safety document generation, and what each contributes to output quality, remain open questions.

### *3.3 Research gaps and proposed system*

Despite the individual maturity of fine-tuning, retrieval augmentation, and multi-agent architectures, two specific gaps emerge from the preceding review. First, the automated generation of daily safety communications in construction remains largely unexplored. Existing LLM applications in the domain focus on information retrieval, compliance checking, and incident analysis, leaving the proactive synthesis of dynamic site data into actionable safety documents insufficiently investigated. Second, the respective contributions of fine-tuning, RAG, and multi-agent orchestration within a unified safety document generation system have not been empirically evaluated, particularly under the deployment constraints—consumer-grade hardware, data privacy requirements— typical of construction organizations. Without controlled comparison under such constraints, practitioners lack evidence-based guidance on which architectural components are necessary for acceptable output quality and which provide marginal improvement.

To address these gaps, this study proposes the Lightweight Multi-agent DSB System (LM-DSB) and evaluates it through controlled comparison across four configurations and expert assessment by construction management professionals. The system design and its rationale are detailed in Section 4; the evaluation methodology and results are presented in Section 5.

# 4 System Development

This section details the design and implementation of LM-DSB. To address the two gaps identified in Section 3.3—the absence of automated DSB generation systems and the lack of empirical evidence on component-level contributions—LM-DSB integrates three components, each addressing specific requirements derived from document analysis (Section 4.1): (1) a hierarchical multi-agent architecture that handles cross-referential dependencies and procedural complexity through task decomposition; (2) a retrieval-augmented generation component that grounds outputs in authoritative sources by leveraging the hierarchical structure of safety documentation; and (3) domain-specific fine-tuning that adapts the base model to specialized terminology and reasoning patterns. The overall architecture is illustrated in Figure 1.

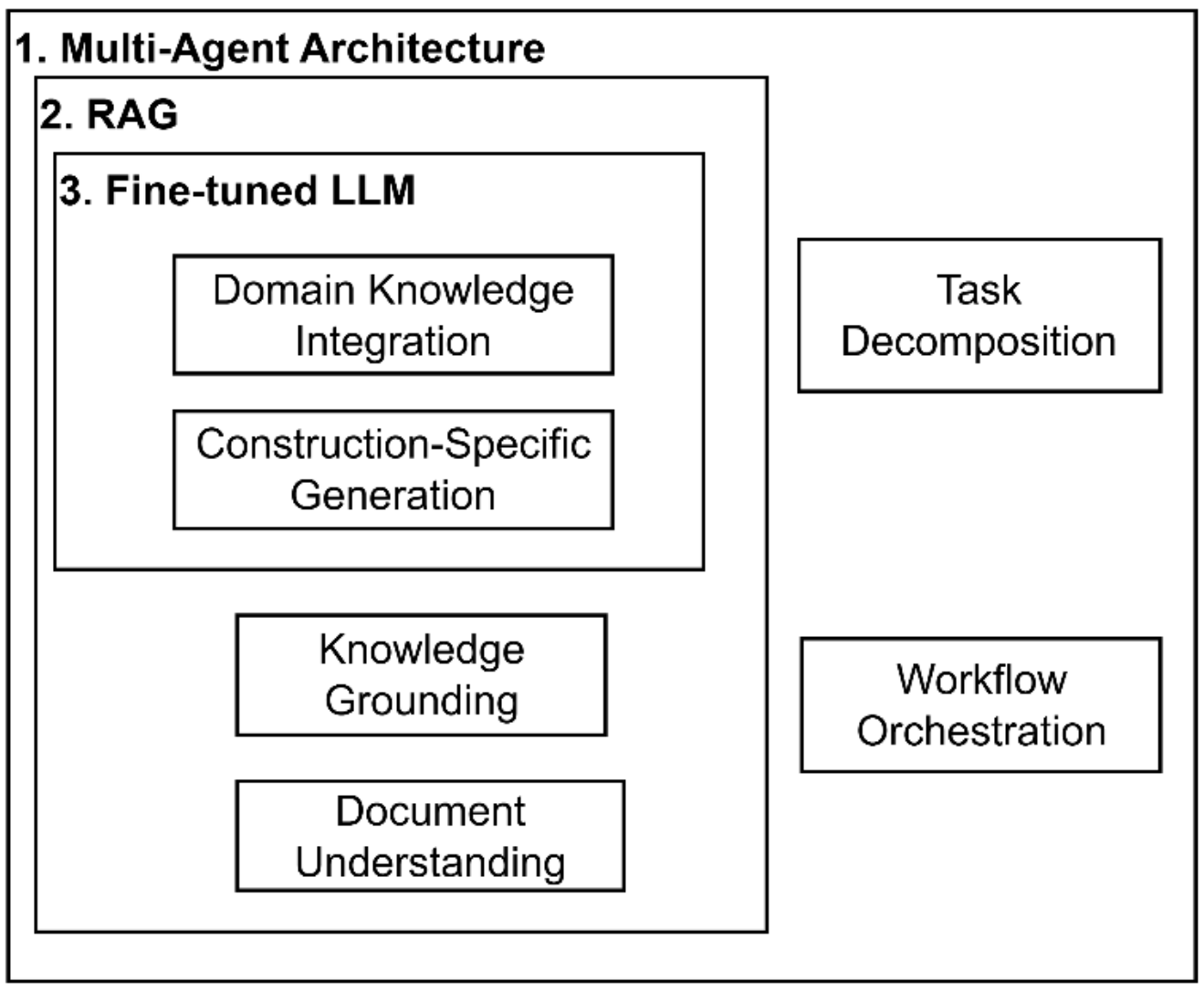

*Figure 1 LM-DSB system's architecture*

The following sections detail this methodology. Section 4.1 analyzes the characteristics of construction safety documentation and derives the design requirements that guided subsequent development. Sections 4.2 through 4.4 then describe how each system component addressed these requirements: multi-agent architecture (4.2), RAG implementation (4.3), and model fine-tuning (4.4).

### *4.1 Data Source*

The proposed methodology begins with the analysis of domain documentation to derive design requirements. Our dataset comprises 37 health, safety, and environment (HSE) documents from a major industrial construction company, categorized into three types: core policy documents (3 documents, 313,958 words) establishing overarching safety frameworks, operational procedures (23 standardized forms, 989 words) covering specific construction activities, and PPE specifications (11 documents, 9,432 words) detailing equipment requirements (Table 1).

*Table 1 Data source information*

| Category | Number of Documents | Pages | Words |
|---|---|---|---|
| Core Policy Documents | 3 | 1,240 | 313,958 |
| Operational Procedures | 23 | 44 | 989 |
| PPE Specifications | 11 | 13 | 9,432 |

Analysis of this corpus reveals three characteristics that directly inform system design. First, the documentation exhibits a hierarchical knowledge structure, ranging from high-level policy frameworks to specific equipment requirements, necessitating retrieval mechanisms capable of operating across multiple granularity levels (Section 4.3). Second, safety protocols demonstrate cross-referential dependencies—generating comprehensive guidance for a single task often requires synthesizing information scattered across multiple documents, motivating a multi-agent architecture for cross-document coordination (Section 4.2). Third, operational procedures follow consistent procedural patterns aligned with Job Hazard Analysis methodology, enabling systematic task decomposition for content generation (Section 4.2). Additionally, the pervasive domain-specific terminology requires adaptation of general-purpose language models through fine-tuning (Section 4.4).

### *4.2 Multi-Agent System Architecture Design*

As identified in Section 4.1, the cross-referential dependencies and procedural patterns inherent in construction safety documentation present challenges that exceed the capacity of a single LLM. Generating comprehensive DSBs requires simultaneously retrieving information from multiple document sources, synthesizing cross-document content, and structuring outputs according to established safety protocols—tasks that benefit from specialized division of labor. This motivates the adoption of a multi-agent system (MAS), defined here as a collection of specialized language model agents that collaborate on interdependent subtasks. Unlike a single-agent approach where one model must handle all responsibilities within limited context windows, a MAS distributes roles among agents optimized for distinct functions such as planning, information retrieval, and content generation.

To operationalize this design, our system employed LangGraph (Zhang et al., 2025), an open-source framework that facilitates the construction of agent-based workflows. LangGraph enabled explicit definition of agent interaction sequences, context persistence across subtasks, and passing of outputs between agents. The following sections detail the MAS design process. Section 4.2.1 explains how we decomposed DSB generation into subtasks by drawing on Job Hazard Analysis principles and assigned specific responsibilities to each agent. Section 4.2.2 then describes the coordination protocol governing agent interactions and information flow.

### 4.2.1 MAS Architecture Design

To address the cross-referential dependencies and procedural patterns identified in Section 4.1, we adopted a hierarchical multi-agent architecture with role assignments informed by the Job Hazard Analysis (JHA) method. JHA is an established method in construction safety management that structures safety analysis into three sequential steps: breaking down a job into component tasks, identifying hazards associated with each task, and defining corresponding preventive measures. This proven method provided a principled basis for decomposing DSB generation into distinct agent responsibilities.

As illustrated in Figure 2, the architecture places a supervisor agent at the top level to coordinate three specialized agents. The planner agent decomposes daily safety scenarios into manageable subtasks, mirroring JHA's first step of task breakdown. The researcher agent identifies relevant hazards, regulatory requirements, and safety protocols from the document knowledge base, corresponding to JHA's hazard identification step. The writer agent synthesizes retrieved information into structured safety communications with preventive guidance, aligning with JHA's final step of defining preventive measures. Each agent is a specialized LLM instance equipped with role-specific prompts optimized for its designated function within the workflow.

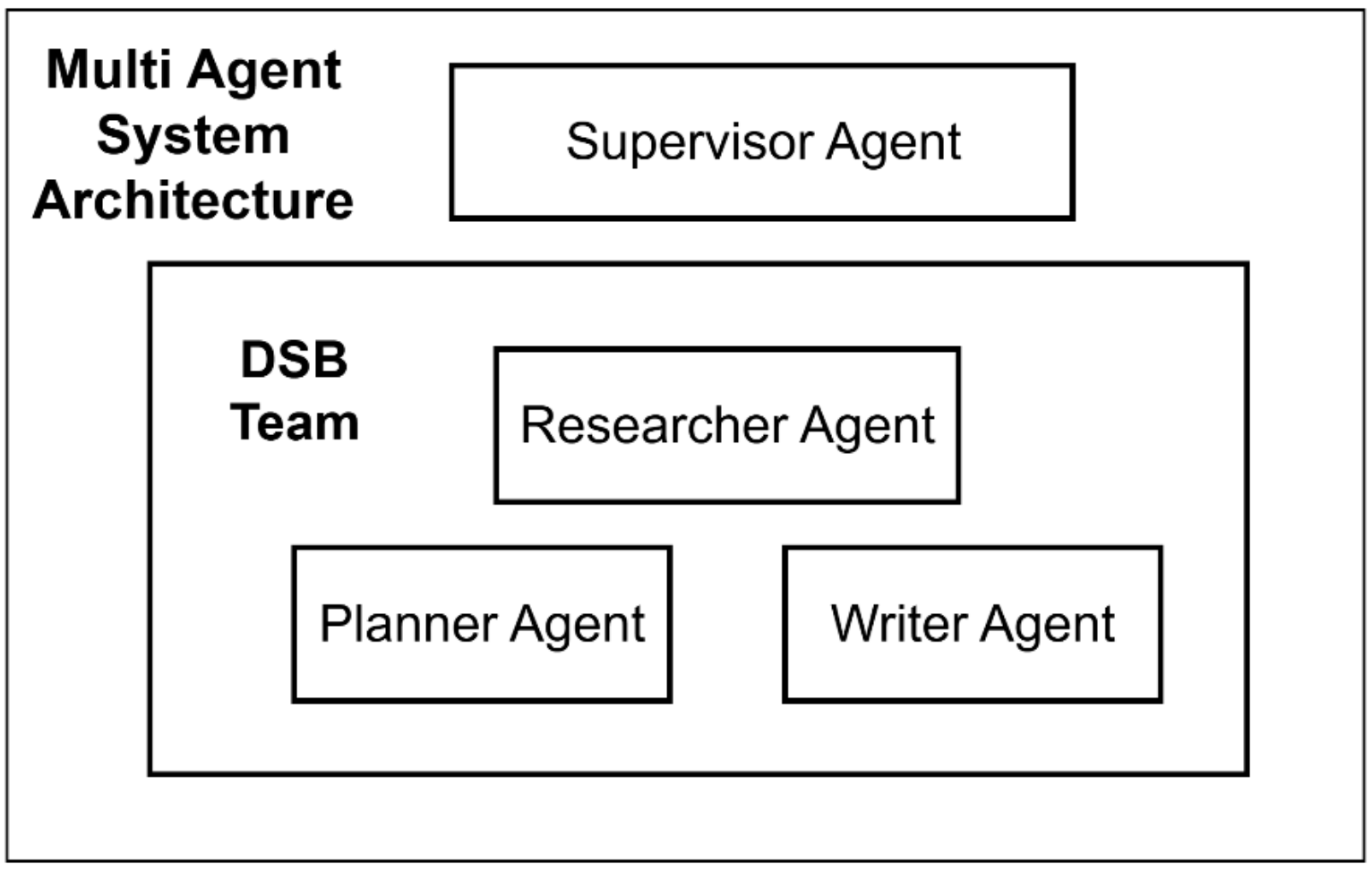

*Figure 2 Proposed hierarchical multi-agent system architecture*

Table 2 summarizes the core prompt design for each agent. The Supervisor agent operates as the workflow coordinator, routing user requests to the appropriate processing pathway. When a DSB generation request is received, the Supervisor delegates sequentially to three specialized agents. The Planner receives real-time context (weather data, project information) and the user-specified safety topic, producing a structured draft covering 11 standard DSB sections and generating targeted research queries for topics requiring regulatory grounding. The Researcher receives these queries along with document chunks retrieved from the vector database, compiling them into structured safety summaries. The Writer then integrates the Planner's structural framework with the Researcher's regulatory content to produce the final DSB.

*Table 2 Agent prompt design specification*

| Agent | Role Definition | Input | Key Instructions | Output |
|---|---|---|---|---|
| Supervisor | Coordinates workflow; routes user requests to appropriate agents or tools | User query | Respond to greetings directly; search HSE documentation for technical queries; delegate DSB creation to the planner-researcher-writer pipeline | Tool call decision or direct response |
| Planner | Generates structured DSB plan with 11 standard sections | Weather data, project information, user-specified topic | Populate sections using available context; generate research queries for Site Specific Communications section to guide the Researcher | JSON with section drafts + research query list |
| Researcher | Synthesizes retrieved safety documentation into structured summaries | Retrieved HSE document chunks (via RAG), Planner's research queries | Compile point-form summaries prioritizing critical safety information; target audience is construction site personnel | Compiled research content with source references |
| Writer | Compiles final DSB by integrating Planner draft with Researcher findings | Planner's section drafts, Researcher's compiled content | Integrate research into Site Specific Communications; ensure concise, professional, and readable output for site workers | Complete DSB document in Markdown |

### 4.2.2 Agent Interaction Flow

Building upon the agent roles defined in Section 4.2.1, we designed a coordination protocol governing information flow between agents. As illustrated in Figure 3, the workflow follows a three-phase sequential structure—planning, research, and writing—orchestrated by a supervisor agent that manages task delegation and context passing.

This sequential design reflects a deliberate architectural choice. Rather than allowing agents to operate in parallel, we enforced strict phase dependencies: the researcher agent cannot begin until the planner agent has identified specific research topics, and the writer agent cannot generate content until research results are available. This constraint ensures that each phase operates on complete inputs from the preceding phase, preventing the fragmented outputs that would result from premature generation. The supervisor agent maintains workflow state and ensures that context accumulated in earlier phases—including the planner's identified hazards and the researcher's retrieved regulations—is systematically passed to downstream agents rather than lost between invocations.

A key integration point occurs during the research phase, where the researcher agent queries the vector database (detailed in Section 4.3) to retrieve relevant HSE documentation. This phase bridges the multi-agent architecture with the RAG component, enabling knowledge-grounded generation in the subsequent writing phase.

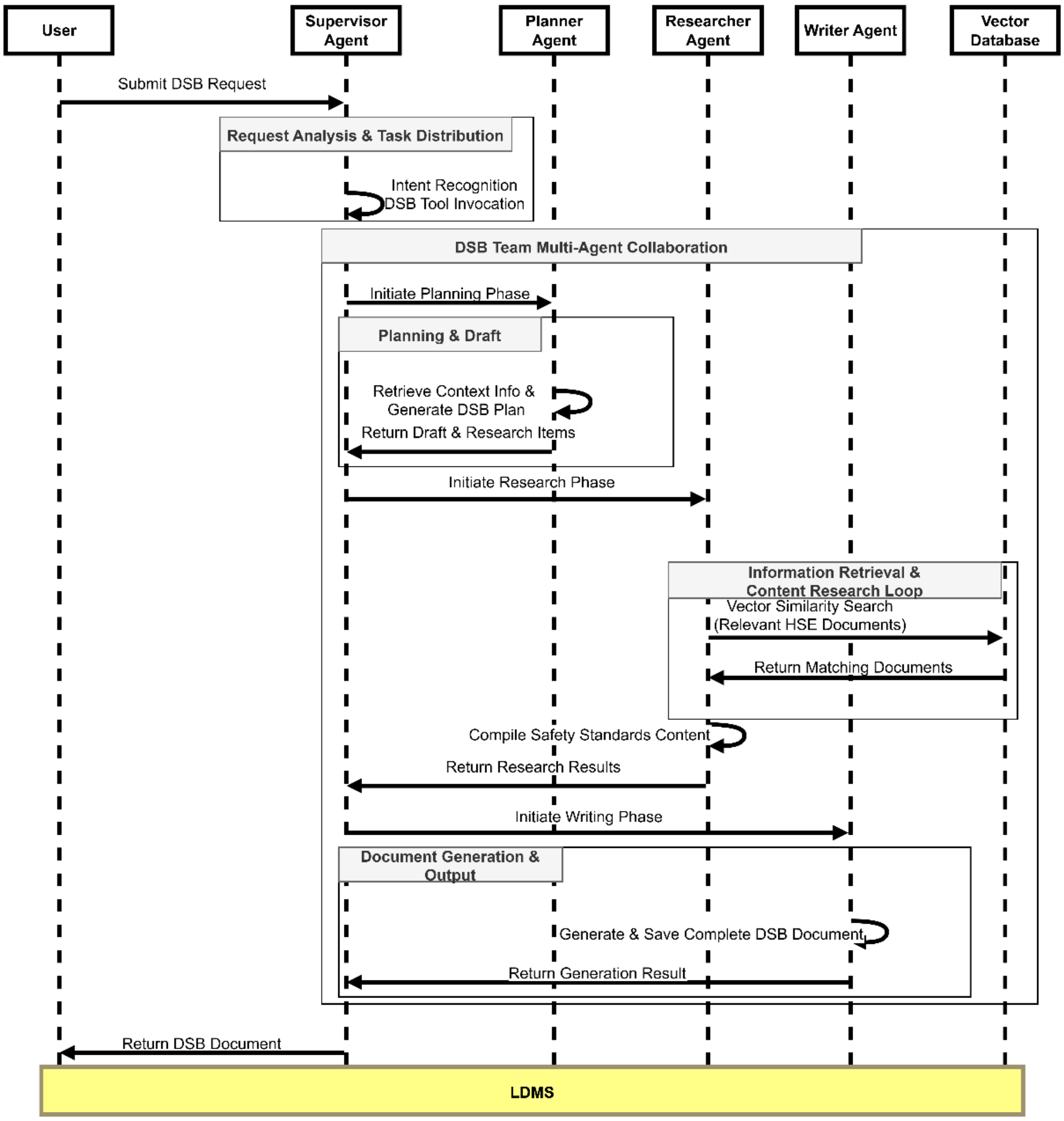

*Figure 3 Interaction sequence flow*

### *4.3 RAG Implementation*

As identified in Section 4.1, the hierarchical structure of construction safety documentation necessitates retrieval mechanisms capable of operating across multiple granularity levels. To address this requirement, we implemented a retrieval-augmented generation (RAG) component (Lewis et al., 2020). that grounds agent outputs in authoritative safety documentation, extending recent applications in construction safety contexts (Prieto et al., 2023; Uhm et al., 2025) through integration with multi-agent task decomposition.

The RAG implementation comprises two elements: a vector database for document storage and a retrieval mechanism integrated with the multi-agent workflow. For the vector database, we processed the 37 HSE documents through a pipeline that segments content into chunks of 1024 tokens with 200-token overlaps, then encodes each chunk using the BGE embedding model (BAAI/bge-large-en-v1.5) to generate 768-dimensional vector representations. These vectors, along with metadata including source file names, page numbers, and document categories, are stored in ChromaDB. Table 3 summarizes the key configuration parameters.

*Table 3 Technical parameter for embedding model*

| Parameter | Value |
|---|---|
| Embedding Model | BAAI/bge-large-en-v1.5 |
| Chunk Size | 1024 tokens |
| Chunk Overlap | 200 tokens |
| Similarity Thresholds | 0.65 |

The retrieval mechanism operates during the research phase described in Section 4.2.2. When the researcher agent receives research topics from the planner agent, it queries the vector database to retrieve the top 3 most relevant document chunks, applying a similarity threshold of 0.65. These parameters were determined through iterative testing on representative safety scenarios—lower thresholds admitted excessive irrelevant content while higher thresholds excluded critical details. Retrieved chunks, together with their source metadata, are then passed to the writer agent, which injects this content into the generation prompt to produce DSBs grounded in actual regulatory language. The preserved metadata enables automatic citation of source documents in the final output (Figure 4), supporting regulatory verification and compliance traceability.

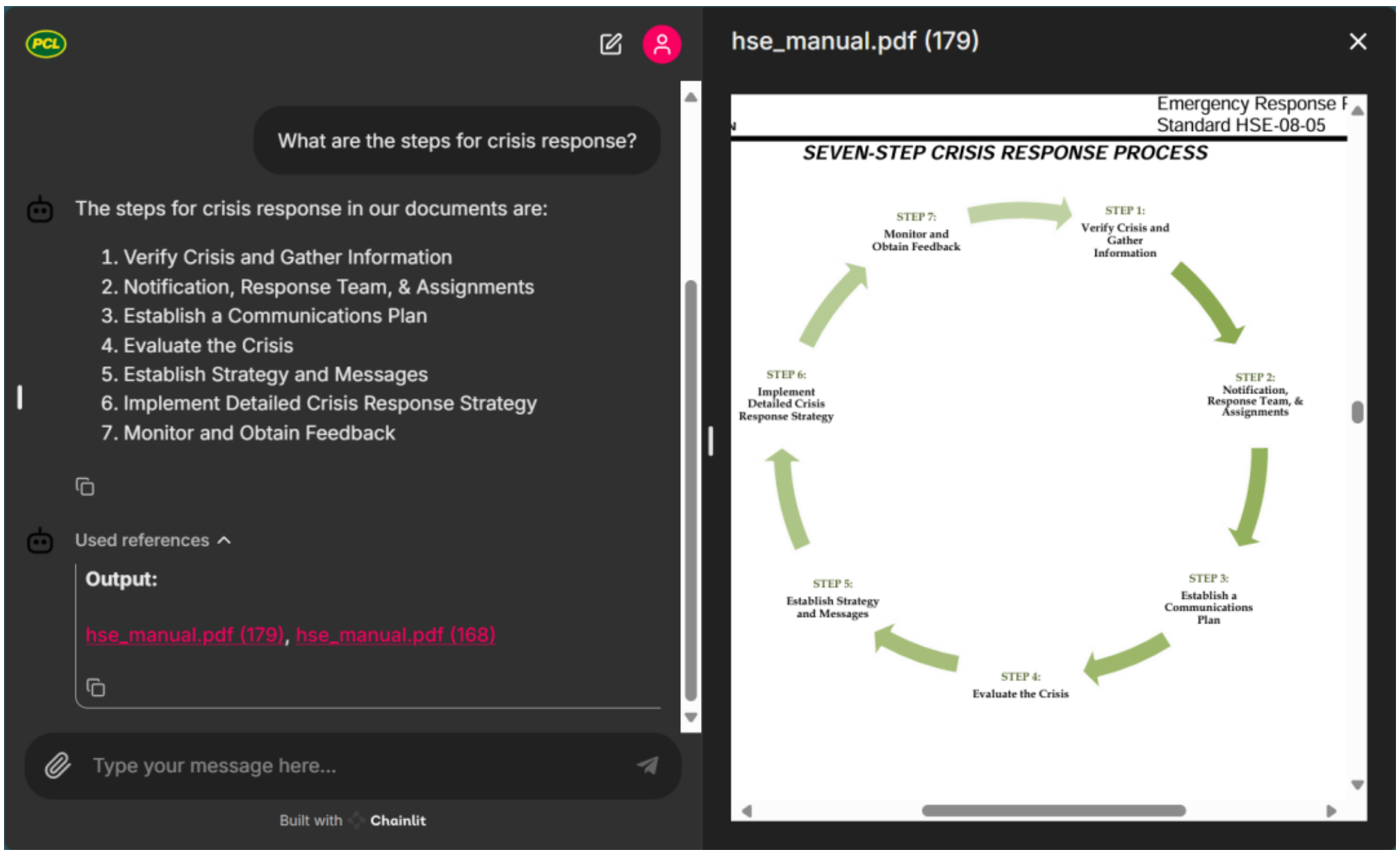

*Figure 4 Example for direct links to source documents*

## *4.4 Model Fine-tuning*

As identified in Section 4.1, the pervasive domain-specific terminology in construction safety documentation requires adaptation of general-purpose language models. While RAG provides access to explicit document content at inference time, fine-tuning enables the model to internalize domain reasoning patterns and terminology. This section describes our base model selection and training data preparation process.

### 4.4.1 Base Model Selection

We selected Llama 3-8B as the base model based on two criteria: local deployability on consumer-grade hardware and sufficient baseline capability for domain adaptation. Table 4 summarizes the candidate models evaluated, with baseline capability measured by the Massive Multitask Language Understanding (MMLU) benchmark, which evaluates model performance across 57 subjects spanning science, engineering, and professional domains.

*Table 4 Selection metrics for base model*

| Model Architecture | Parameter Count | Context Length | MMLU (5-shot) |
|---|---|---|---|
| Meta LLaMA-3 | 8B | 8,192 tokens | 66.60% |
| Google Gemma | 7B | 8,192 tokens | 64.30% |
| Mistral-8x7B MoE | 46.7B (13B/token) | 32,768 tokens | 70.60% |
| Microsoft Phi-3-mini | 3.8B | 131,072 tokens | 68.10% |

Among lightweight options (≤8B parameters), Llama 3-8B offered the strongest baseline performance (MMLU: 66.6%) while remaining deployable on standard workstation GPUs (12-16GB VRAM) commonly available in construction offices. Although Phi-3-mini provides a larger context window and Mistral-8x7B achieves higher MMLU scores, the former's smaller parameter count limits fine-tuning capacity, while the latter's 46.7B parameters exceed consumer hardware constraints. Llama 3-8B thus represented the optimal balance between accessibility and adaptability for our domain-specific application.

#### 4.4.2 Data processing

To generate fine-tuning data, we developed a pipeline that transforms unstructured HSE documents into structured training examples. The pipeline employs the "unstructured" library for content extraction, automatically classifying content into text, tables, and images (Table 5). Text segments are processed directly, while tables and images are summarized using GPT-4 and GPT-4-Vision respectively to generate textual descriptions suitable for training.

*Table 5 Elements in our data processing*

| Content | Processing Method | Output Format | Model | Content |
|---|---|---|---|---|

| Type | | | Used | Type |
|---|---|---|---|---|
| Text | Composite Element Processing | Structured Text | - | Text |
| Tables | GPT-4 Table Summarization | JSON Structure | GPT-4 | Tables |
| Images | Vision-based Processing | Base64 + Summary | GPT-4-Vision | Images |
| Metadata | Automated Extraction | JSON Fields | - | Metadata |

Based on different types of content we have, we have 2 different processes shown in Figure 5. We implement text segmentation using the unstructured library with parameters: 4000-character maximum length, 3800-character new segment threshold, and 2000-character minimum for segment consolidation. These parameters balance segment size with contextual coherence:

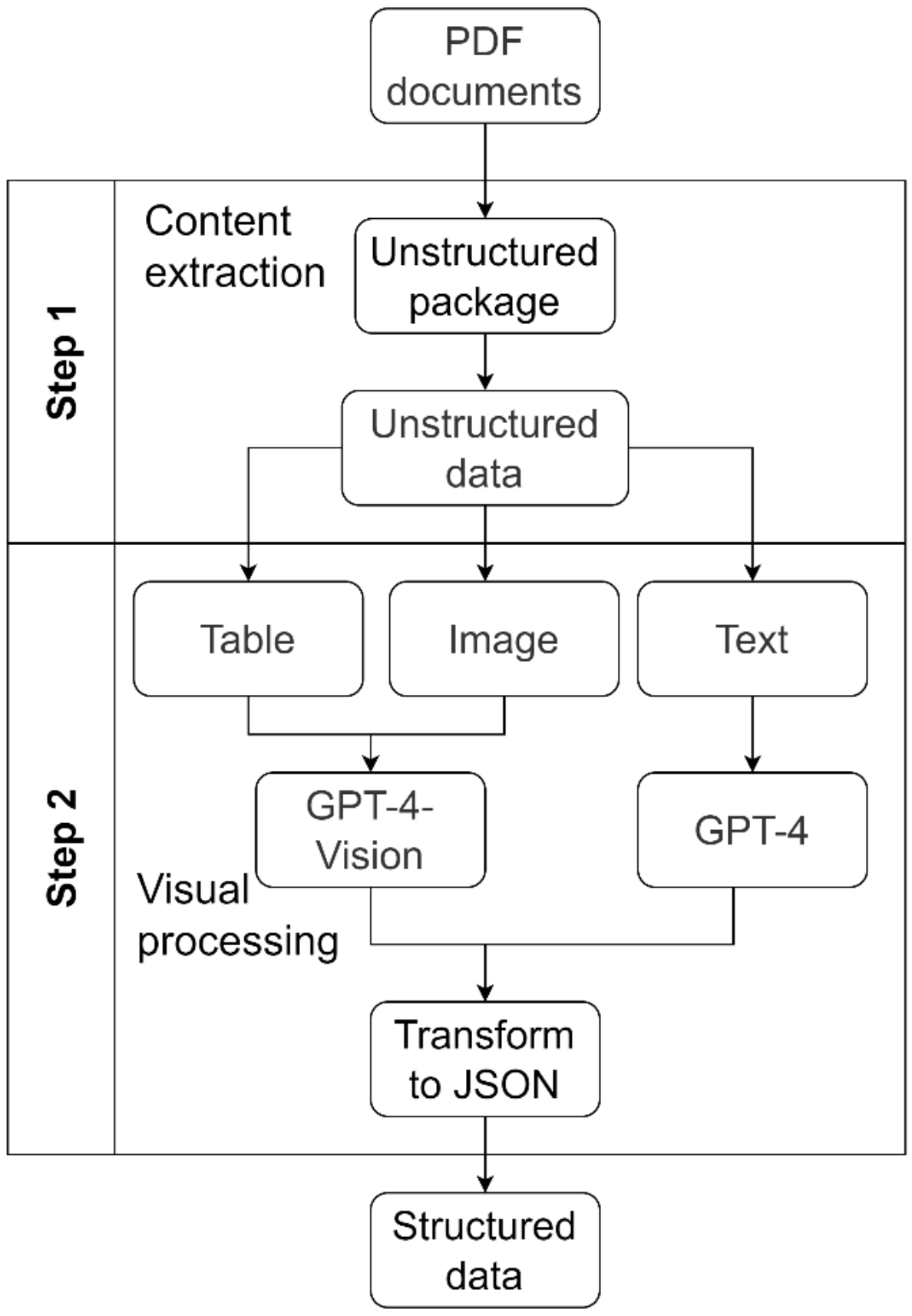

*Figure 5 Data processing pipeline*

### 4.4.3 Q-A pair generation

A key methodological challenge in adapting LLMs for construction safety applications is the absence of publicly available domain-specific training datasets. To address this gap, we developed an automated pipeline that transforms the processed HSE documents from Section 4.4.2 into structured question-answer pairs suitable for supervised fine-tuning.

The pipeline combines text segments in groups of three to generate questions requiring cross-referential information synthesis—directly reflecting the cross-referential dependencies identified in Section 4.1. Each segment combination produces six question-answer pairs using GPT-4 with temperature set to 0 for consistency. The resulting training data follows a structured format (Table 6) that includes not only prompts and completions but also source chunk identifiers, enabling traceability back to original HSE documentation. This approach generated 10,837 training examples from our 37-document corpus, demonstrating that proprietary construction safety documentation can be systematically converted into effective LLM training resources.

*Table 6 Data structure for model fine-tuning*

| Field | Format | Description | Example |
|---|---|---|---|
| Prompt | Text String | Contains the system instruction, question, and context | "You are an expert... Question: What are safety protocols? Context: Safety guidelines..." |
| Completion | Text String | Contains the formatted answer and source reference | "Here is my answer: Safety protocols include... Here is the ID: 125" |
| ChunkIDs | String | Unique identifiers for source text chunks | "125, 126, 127" |
| ChunkTexts | Text | Source material used for answer generation | "Section 1.1: Safety protocols..." |

#### 4.4.4 Model Fine-tuning

We employed supervised fine-tuning (SFT) with LoRA (Low-Rank Adaptation) to adapt Llama 3-8B to the construction safety domain (Hu et al., 2021; Ouyang et al., 2022). LoRA reduces trainable parameters by approximately 99.8% through low-rank matrix decomposition, enabling fine-tuning on consumer-grade GPUs (12-16GB VRAM) without sacrificing adaptation quality. Combined with 4-bit quantization, this configuration allows domain adaptation on the same workstation hardware used for BIM modeling in construction offices. Table 7 details the implementation parameters;

the effectiveness of this configuration is validated in Section 5.

*Table 7 Fine-tuning parameters*

| Parameter | Value |
|---|---|
| Base Model | LLaMA-3-8B-Instruct |
| Quantization | 4-bit |
| Fine-tuning Method | LoRA |
| Batch Size | 2 |
| Learning Rate | 5.00E-05 |
| Training Epochs | 3 |
| Gradient Accumulation | 4 steps |
| Scheduler | Cosine with 0.1 warmup |
| LoRA+ Ratio | 16 |
| Max Gradient Norm | 1 |
| Max Samples | 500 |

## 5 Validation of the concept and discussion

This section evaluates the feasibility and effectiveness of the proposed multi-agent RAG system for automated construction safety documentation generation. The validation examines whether the integrated approach—combining retrieval augmentation, domain-specific fine-tuning, and multi-agent orchestration—can produce safety bulletins that meet professional quality standards, while simultaneously assessing whether architectural innovation can compensate for limited model capacity in lightweight LLMs (≤8B parameters). We employed a two-phase validation strategy: Phase 1 establishes technical performance benchmarks through comparative quantitative analysis; Phase 2 assesses practical viability through expert evaluation and operational case studies. All experiments were conducted on consumer-grade hardware (Intel Core i5-13600KF, NVIDIA RTX 4070 Ti 12GB VRAM, 32GB RAM).

### *5.1 Comparative Quantitative Analysis*

We constructed a controlled comparative experiment evaluating four distinct configurations—Llama-Base, Llama-FT (fine-tuned), Llama-RAG (retrieval-augmented generation), and LM-DSB (Lightweight Multi-agent Daily Safety Bulletin)—across three standardized construction safety scenarios to isolate the individual performance contributions of fine-tuning, retrieval augmentation, and agent orchestration. Llama-Base employs unmodified Llama 3-8B to establish baseline capability. Llama-FT incorporates domain-specific fine-tuning without RAG to assess whether specialized training alone suffices. Llama-RAG integrates BGE embeddings with a vector database and a single LLM instance to quantify retrieval augmentation benefits. LM-DSB implements the proposed hierarchical multi-agent architecture with planner-researcher-writer coordination.

Each configuration was evaluated across three standardized construction safety scenarios: (1) tower crane and mobile crane operations in extreme cold conditions, (2) trenching and excavation combined with confined space entry, and (3) hazardous energy control with compressed gas safety. All four configurations processed identical inputs for each scenario. The input to each configuration consists of a single natural language instruction requesting DSB generation for a specific topic, such as: *"Generate a Daily Safety Bulletin about tower cranes and mobile crane operations, focusing on key risks and preventive measures."* The system then autonomously retrieves environmental context (e.g., real-time weather data), identifies relevant regulatory requirements, and generates a complete DSB without additional user input. Detailed scenario descriptions, sample inputs, and representative outputs are presented in

Section 5.2.

Ground truth DSBs (515–620 words) were developed by domain experts with comprehensive hazard coverage and regulatory compliance. Generated outputs were evaluated using four complementary NLP metrics—TF-IDF similarity, BLEU score, ROUGE-1 F1, and ROUGE-2 F1—that collectively measure lexical alignment, phrase precision, semantic coherence, and content completeness. Additionally, generation length ratio (output words / ground truth words) assessed document completeness.

Quantitative results revealed that RAG-enhanced configurations (Llama-RAG, LM-DSB) substantially outperformed non-RAG alternatives (Llama-Base, Llama-FT) across all four metrics. The most pronounced difference appeared in bigram coherence, where RAG configurations achieved 4–5× improvements over non-RAG alternatives (ROUGE-2: LM-DSB = 0.41, Llama-RAG = 0.38 vs. Llama-Base = 0.08, Llama-FT = 0.07), confirming that knowledge grounding enables phrase-level factual coherence unattainable through model capability alone.

The multi-agent architecture (LM-DSB) provided substantial advantages over single-agent RAG (Llama-RAG), with consistent improvements across all metrics (TF-IDF: 0.70 vs. 0.63; BLEU: 0.39 vs. 0.27; ROUGE-1: 0.66 vs. 0.48). Most critically, LM-DSB achieved near-complete document generation (length ratio: 1.01) compared to Llama-RAG's 0.39—indicating that single-agent RAG produced only about one-third of the expected content. This completeness differential demonstrates that agent orchestration enables small-parameter models to overcome individual capacity limitations through specialized task decomposition, allowing the 8B model to produce outputs comparable in length and structure to expert-authored documents.

The Llama-FT result provides important insight into both the role of fine-tuning and the limitations of the evaluation metrics. As a standalone configuration without RAG, Llama-FT scored below Llama-Base across all four metrics (TF-IDF: 0.42 vs. 0.49; BLEU: 0.08 vs. 0.11; ROUGE-1: 0.36 vs. 0.41; ROUGE-2: 0.07 vs. 0.08), consistent with findings that fine-tuning LLMs on narrow domain knowledge can degrade general generative capabilities, particularly in smaller-parameter models (Gekhman et al., 2024).

However, the evaluation metrics employed—TF-IDF, BLEU, and ROUGE—measure lexical overlap with specific ground truth documents rather than domain-appropriate expression. Fine-tuning optimizes the model for construction safety terminology and reasoning patterns, which may produce functionally equivalent but lexically divergent outputs, causing these metrics to underestimate fine-tuning's practical contribution.

Within LM-DSB, fine-tuning serves a complementary rather than primary role: RAG provides factual grounding, multi-agent decomposition manages task complexity, and fine-tuning enhances fluency with domain-specific conventions. LM-DSB outperformed Llama-RAG across all metrics (TF-IDF +11%, BLEU +44%, ROUGE-1 +38%), but this gap reflects the combined contribution of multi-agent orchestration and the fine-tuned backbone; the current experimental design does not fully isolate their individual effects. Given that the most substantial gain—the length ratio increase from 0.39 to 1.01—is primarily attributable to agent task decomposition, we consider orchestration the dominant contributor, with fine-tuning providing supplementary domain adaptation.

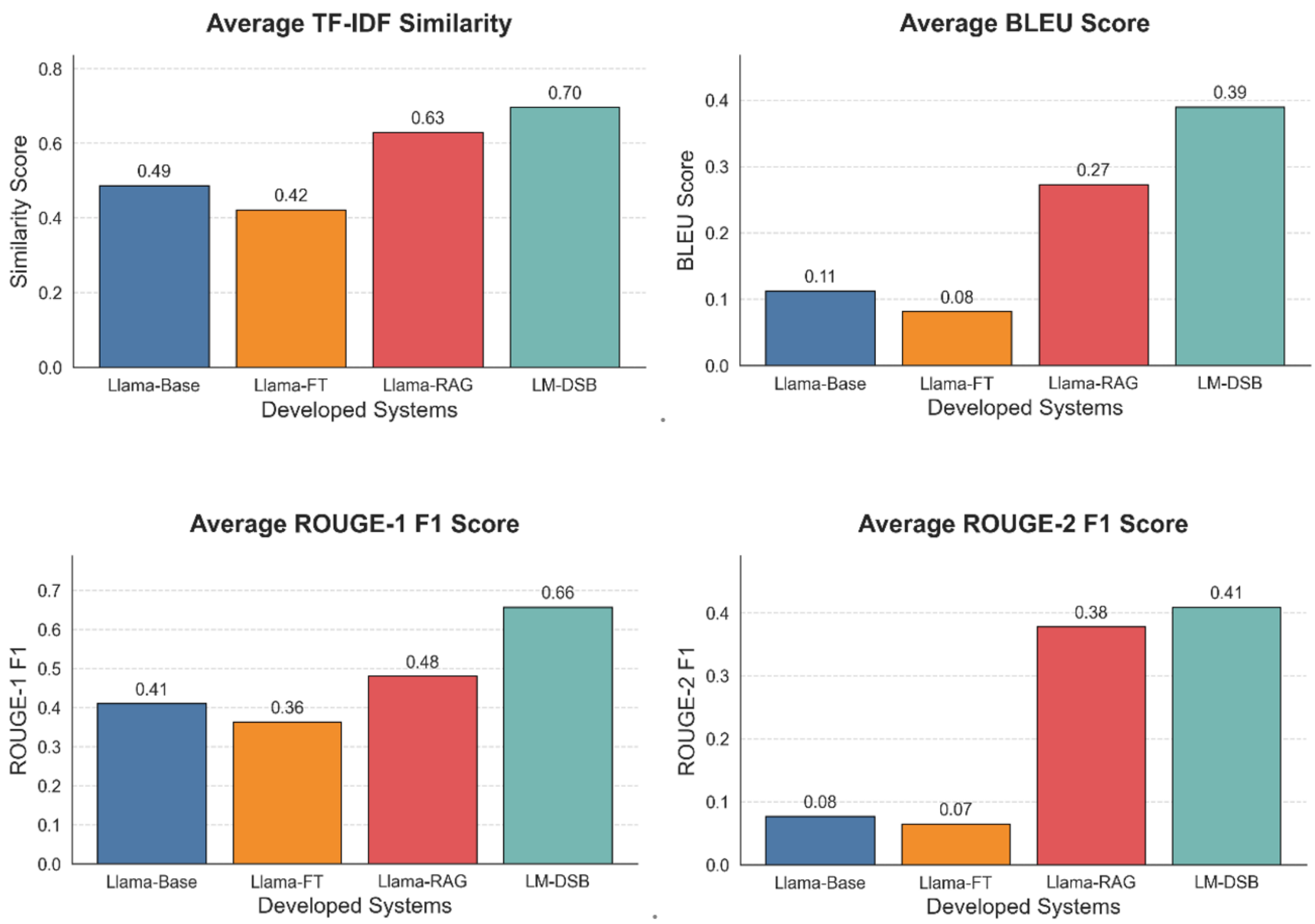

*Figure 6 Performance comparison across 4 NLP metrics*

*Table 8 Quantitative performance comparison*

| System | TF-IDF | BLEU | ROUGE-1 | ROUGE-2 |
|---|---|---|---|---|
| **LM-DSB** | **0.6960** | **0.3898** | **0.6570** | **0.4086** |
| Llama-RAG | 0.6292 | 0.2731 | 0.4804 | 0.3781 |
| Llama-FT | 0.4222 | 0.0814 | 0.3631 | 0.0651 |
| Llama-Base | 0.4870 | 0.1124 | 0.4099 | 0.0767 |

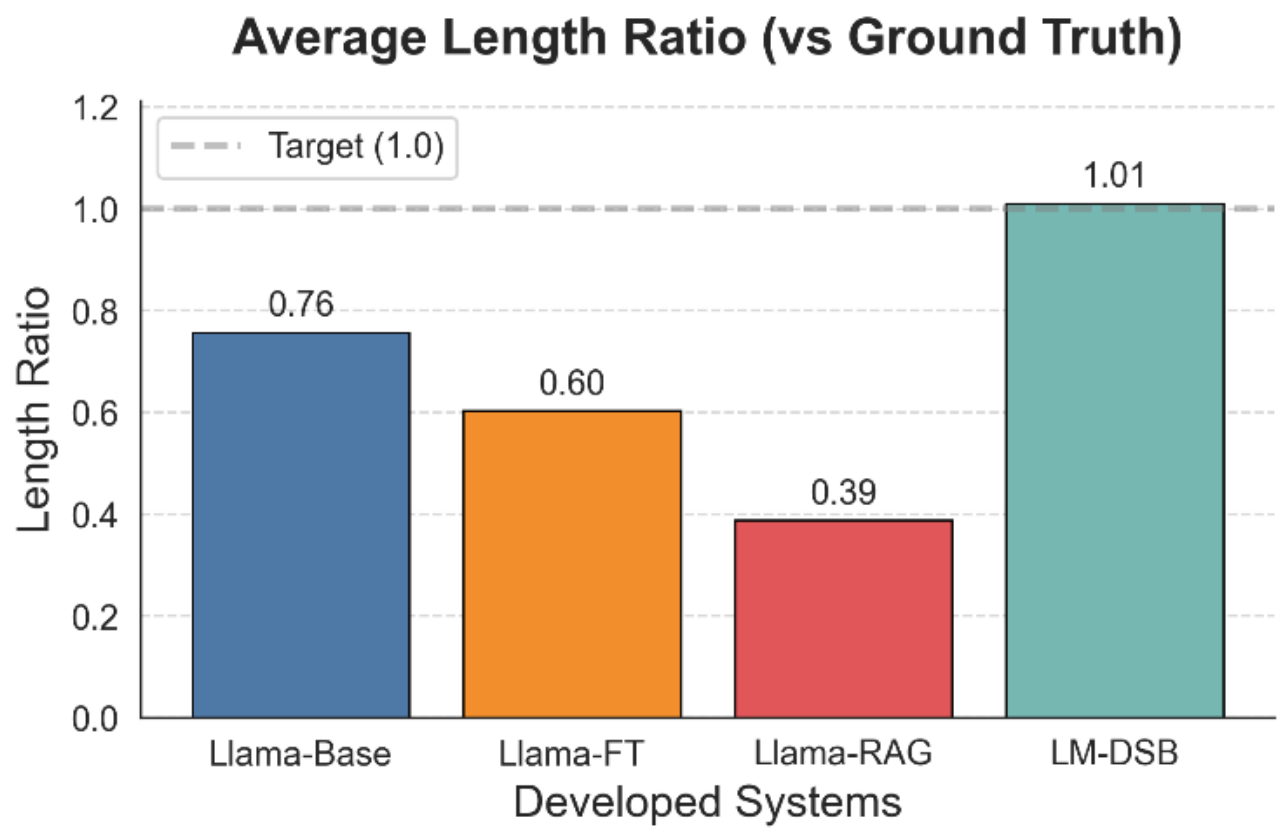

*Figure 7 Average Length Ratio Comparison*

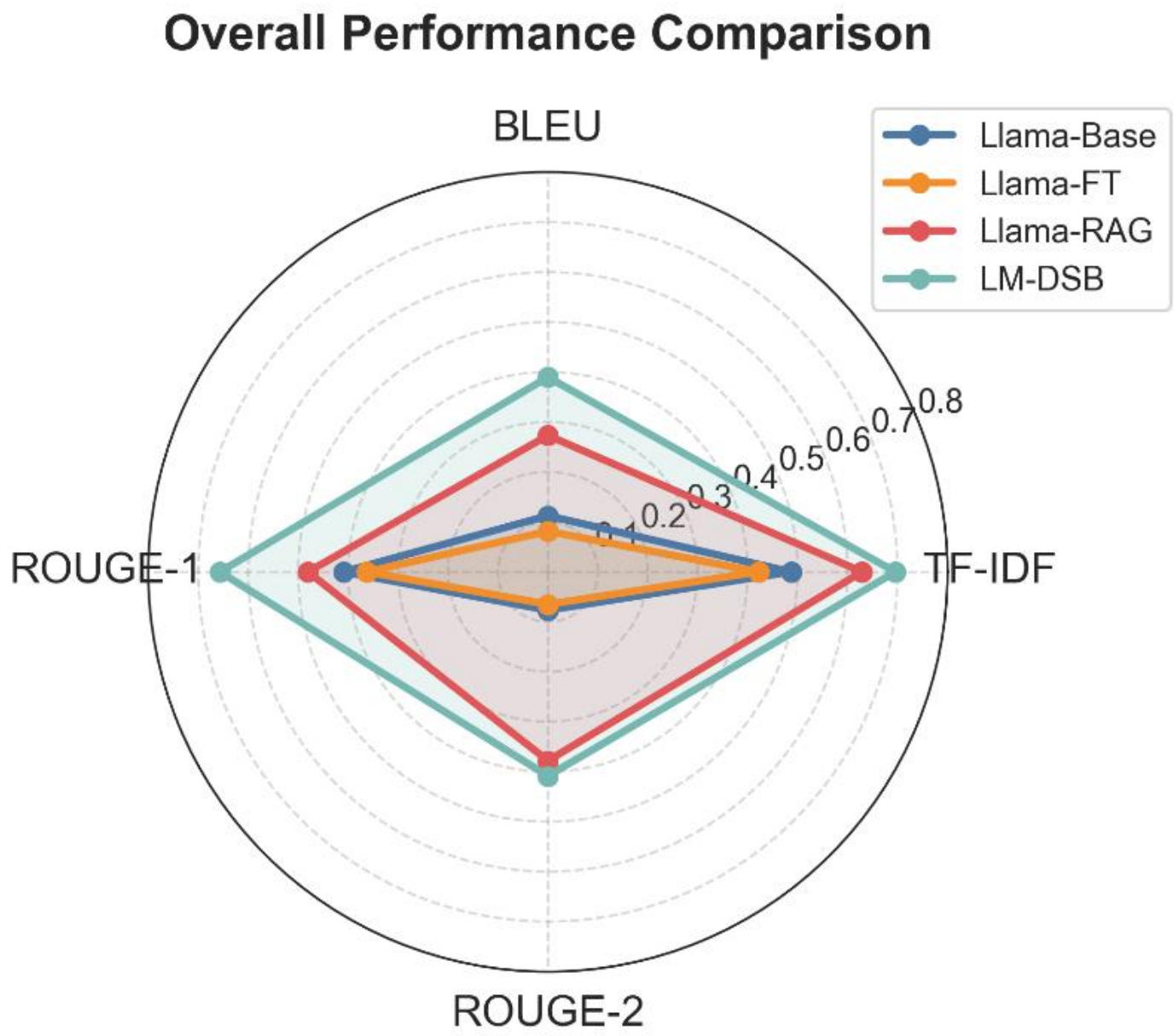

*Figure 8 Overall Performance Comparison*

These comparative results establish LM-DSB as the best-performing configuration among the four evaluated, demonstrating both the necessity of retrieval augmentation for factual accuracy and the contribution of multi-agent orchestration to generation completeness in lightweight models. The following section consequently evaluates this configuration's practical viability through human assessment and operational case studies.

### *5.2 Expert Evaluation and Operational Assessment*

To verify whether quantitative performance corresponds to human-perceived quality, construction management professionals (N=40) evaluated system-generated DSBs across five dimensions: (1) factual accuracy & groundedness, (2) completeness & comprehensiveness, (3) comprehensibility & practical usability, (4) actionability & behavioral guidance, and (5) regulatory compliance & standards alignment. Evaluators rated each dimension on a 5-point Likert scale (1=Poor, 5=Excellent) after reviewing three system-generated DSBs from the scenarios in Section 5.1.

The multi-agent system achieved robust performance across all dimensions (mean score M = 3.50–4.20 on the 5-point scale), with highest ratings for Regulatory Compliance (M = 4.20, SD = 0.65) and Factual Accuracy (M = 4.10, SD = 0.55), where SD denotes standard deviation. These results confirm that RAG-based knowledge grounding successfully maintains information integrity and retrieves relevant safety codes. Mid-range scores for Completeness (M = 3.98, SD = 0.62) and Comprehensibility (M = 3.77, SD = 0.86) indicate adequate coverage and accessibility for site personnel. Actionability received the lowest rating (M = 3.50, SD = 0.91), suggesting refinement opportunities in procedural specificity and responsibility assignment. The small standard deviations for top-performing dimensions indicate strong evaluator consensus, while larger variation in Actionability reflects diverse expectations for procedural detail.

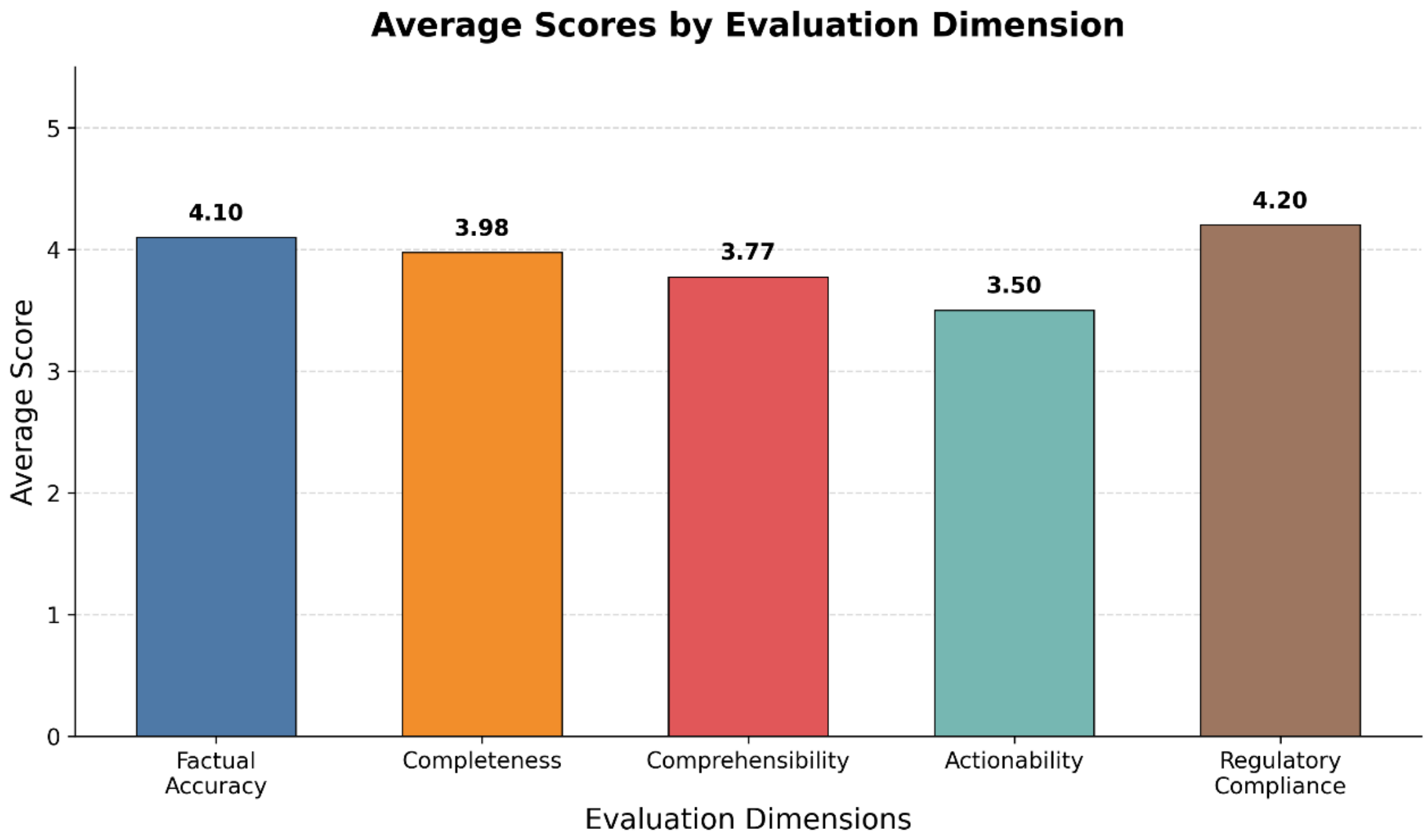

*Figure 9 Expert Evaluation Dimension Scores*

*Table 9 Expert evaluation results (N=40)*

| Dimension | Mean | Median | SD | Min | Max |
|---|---|---|---|---|---|
| **Factual Accuracy & Groundedness** | 4.10 | 4.00 | 0.55 | 3 | 5 |
| **Completeness & Comprehensiveness** | 3.98 | 4.00 | 0.62 | 2 | 5 |
| **Comprehensibility & Practical Usability** | 3.77 | 4.00 | 0.86 | 2 | 5 |
| **Actionability & Behavioral Guidance** | 3.50 | 4.00 | 0.91 | 2 | 5 |
| **Regulatory Compliance & Standards Alignment** | 4.20 | 4.00 | 0.65 | 3 | 5 |

These results substantiate that lightweight LLM-generated output achieves practical acceptability from domain professionals' perspective, transcending mere metric-based performance. The robust Regulatory Compliance and Factual Accuracy scores validate the system's core value proposition: generating safety documentation that maintains authoritative grounding while reducing manual preparation time.

Three case studies examined system behavior in complex multi-hazard scenarios requiring integration of real-time environmental data, regulatory requirements, and context-specific safety protocols. Case 1 (crane operations in -25°C; Figure 10) demonstrated successful integration of weather data with regulatory requirements, automatically incorporating cold-weather protocols and localized hazards. However, corpus-dependent blind spots emerged where missing documents led to critical omissions in specialized technical areas. Case 2 (trenching with confined spaces; Figure 11) exposed the 8B model's inferential reasoning limitations: the system failed to synthesize implicit connections across multiple risk factors (e.g., trenching + cold weather + diesel heating → carbon monoxide buildup) despite having relevant documentation for each component. Case 3 (hazardous energy control; Figure 12) showcased domain fine-tuning benefits through accurate extrapolation of cold-weather equipment risks despite sparse documentation, yet revealed persistent temporal reasoning deficits with vague timeframes rather than specific intervals.

```
**Site Specific Communications**

### Safety Guidelines for Tower Cranes and Mobile Cranes Operations

#### **Specific Hazards Associated with Tower Cranes**

Tower cranes are complex machines that require careful operation to ensure safety. Some
specific hazards associated with tower crane operations include:

* **Overturning**: The crane can topple over if not properly secured or if the load is
too heavy.
* **Swinging**: The crane's boom can swing out of control, causing damage or injury.
* **Entanglement**: Workers can become entangled in the crane's rigging or other
equipment.
* **Electrical hazards**: The crane's electrical systems can pose a shock hazard if not
properly maintained.

#### **Safely Managing Mobile Crane Operations in Cold Weather Conditions**

Mobile cranes can be safely operated in cold weather conditions like Edmonton's current
-25°C temperature by following these guidelines:

* **Verify cold weather ratings**: Ensure that the crane manufacturer provides cold
weather ratings for the specific equipment being used. These ratings will indicate the
maximum loads and speeds allowed under cold weather conditions.
* **Derate crane capacities**: If the manufacturer does not provide cold weather
ratings, derate the crane's capacities to ensure safe operation.
* **Use proper rigging**: Ensure that all rigging is properly secured and rated for the
load being lifted. Consider using heavier-duty rigging in cold weather conditions.
* **Monitor wind speeds**: Monitor wind speeds regularly to avoid operating the crane in
high winds, which can cause instability and increase the risk of overturning or swinging.

#### **Regulations and Guidelines for Operating Tower Cranes and Mobile Cranes in
Alberta, Canada**

In Alberta, Canada, the operation of tower cranes and mobile cranes is regulated by
various laws and guidelines. Some key regulations include:
```

*Figure 10 LM-DSB response for Case 1: Tower crane and mobile crane operations in extreme cold conditions*

**Site Specific Communications**
Trenching and Excavation Safety Requirements Combined with Confined Space Entry

### Trenching and Excavation Safety Requirements in Edmonton
The City of Edmonton requires that all trenching and excavation projects comply with the Alberta Occupational Health and Safety Code. This includes ensuring that all trenches are properly shored and protected from collapse, and that workers enter and exit the trench safely.

### Regulations for Excavation Site Supervision in Alberta
According to the Alberta Occupational Health and Safety Code, excavations must be supervised by a competent person who has knowledge of the hazards associated with excavation work. This includes monitoring weather conditions, ensuring that trenches are properly shored, and conducting regular inspections to ensure that the site is safe for workers.

### Confined Space Entry Procedures for Excavations in Cold Weather Conditions
When working in cold weather conditions, it's essential to take extra precautions when entering confined spaces. This includes ensuring that all equipment is functioning correctly, wearing appropriate personal protective equipment (PPE), and following established procedures for entry and exit.

### Safety Protocols for Workers Entering Trenches with Temperatures Below -25°C
When temperatures drop below -25°C, workers must take extra precautions when entering trenches. This includes ensuring that all equipment is functioning correctly, wearing appropriate PPE, and following established procedures for entry and exit. Additionally, workers should be trained on the hazards associated with working in cold weather conditions.

### Best Practices for Communication between Team Members during Confined Space Entries
Effective communication is critical when performing confined space entries. This includes ensuring that all team members are aware of their roles and responsibilities, communicating clearly about any hazards or risks, and establishing a system for monitoring the work site.

*Figure 11 LM-DSB response for Case 2: Trenching and excavation safety combined with confined space entry*

### Common Hazards Associated with Hazardous Energy Control

Some common hazards associated with hazardous energy control include:
* Electrical shock or electrocution from malfunctioning equipment or poor maintenance
* Crushing or pinching injuries from machinery or equipment that is not properly shut off or secured
* Fire or explosion from ignition sources such as sparks or hot surfaces
* Chemical exposure from leaking or ruptured compressed gas cylinders

### Safe Handling Procedures for Compressed Gas Cylinders

Safe handling procedures for compressed gas cylinders include:
* Ensuring cylinders are stored in a well-ventilated area, away from heat sources and flames
* Keeping cylinders upright and secure to prevent accidents during transport or storage
* Regularly inspecting cylinders for signs of damage or wear, such as dents, rust, or corrosion
* Labeling and identifying compressed gas cylinders with their contents and hazards

### Training Workers on Compressed Gas Safety

Workers can be trained on compressed gas safety by:
* Providing classroom instruction on the safe handling, storage, and use of compressed gases
* Conducting on-the-job training sessions to demonstrate proper procedures for handling compressed gas cylinders
* Encouraging workers to report any concerns or incidents related to compressed gas safety
* Regularly reviewing and updating emergency response plans in case of a compressed gas leak or other incident

*Figure 12 LM-DSB response for Case 3: Hazardous energy control and compressed gas safety*

Collectively, these cases reveal consistent patterns: the system excels at retrieving and adapting explicit regulatory content and performs limited domain-specific extrapolation but encounters systematic failures in complex inferential reasoning and implicit knowledge synthesis. These findings confirm that architectural innovation enables lightweight models to handle document-grounded generation tasks effectively, while also delineating the boundaries where parameter limitations constrain performance.

## 6 Conclusion

This study demonstrated that lightweight language models (≤8B parameters) can produce construction safety documentation of acceptable quality when enhanced with multi-agent architecture and retrieval-augmented generation. The proposed LM-DSB consistently outperformed all alternative configurations across four NLP metrics, and—most critically—achieved near-complete document generation (length ratio: 1.01) where single-agent RAG produced only partial outputs (0.39). Expert evaluations confirmed practical acceptability across quality dimensions (M = 3.50–4.20). These results establish the feasibility of automated DSB generation on consumer-grade hardware, suggesting that multi-agent task decomposition and domain-specific fine-tuning offer a viable path toward accessible, locally deployable safety documentation tools for resource-constrained construction organizations.

Beyond technical validation, this study contributes a practical multi-agent architecture designed specifically for construction safety documentation. The hierarchical planner-researcher-writer structure, informed by established job hazard analysis principles, demonstrates how domain knowledge about construction safety workflows can guide agent role design and interaction patterns. While the current implementation targets

DSB generation, the underlying architectural design—task decomposition, regulatory retrieval, and structured content synthesis—may be adaptable to related construction documentation tasks such as work plans, inspection checklists, and training materials, though further investigation is needed to confirm transferability.

Several limitations constrain the current findings. The knowledge base derives from a single organization's HSE documentation, potentially limiting generalizability across diverse construction contexts. The Llama 3-8B foundation restricts evaluation to English-language documentation. Validation occurred in controlled experimental conditions rather than longitudinal field deployment, and the 8B model exhibited inferential reasoning limitations when synthesizing implicit connections across multiple risk factors.

Future research should address these limitations through several directions: developing reasoning-aware RAG techniques to enhance multi-factor synthesis capabilities; integrating multimodal inputs (site photographs, speech commands) for more intuitive operational use; investigating workflow integration mechanisms that minimize adoption friction; and expanding evaluation to diverse organizational contexts, multiple languages, and extended deployment periods. Additionally, adapting the proposed multi-agent architecture to other construction document types, such as work plans, inspection checklists, and training materials, would further validate its generalizability within the domain.